# A Review of Mechanistic Models of Event Comprehension

Tan T. Nguyen

Department of Psychological and Brain Sciences, Washington University in St. Louis, St. Louis, MO, USA

Corresponding author:

Tan T. Nguyen: n.tan@wustl.edu

# Contents

# 1. Abstract

This review examines theoretical assumptions and computational models of event comprehension, tracing the evolution from discourse comprehension theories to contemporary event cognition frameworks. The review covers key discourse comprehension accounts, including Construction-Integration, Event Indexing, Causal Network, and Resonance models, highlighting their contributions to understanding cognitive processes in comprehension. I then discuss contemporary theoretical frameworks of event comprehension, including Event Segmentation Theory (Zacks et al., 2007), the Event Horizon Model (Radvansky & Zacks, 2014), and Hierarchical Generative Framework (Kuperberg, 2021), which emphasize prediction, causality, and multilevel representations in event understanding. Building on these theories, I evaluate five computational models of event comprehension: REPRISE (Butz et al., 2019), Structured Event Memory (SEM; Franklin et al., 2020), the Lu model (Lu et al., 2022), the Gumbsch model (Gumbsch et al., 2022), and the Elman and McRae model (2019). The analysis focuses on their approaches to hierarchical processing, prediction mechanisms, and representation learning. Key themes that emerge include the use of hierarchical structures as inductive biases, the importance of prediction in comprehension, and diverse strategies for learning event dynamics. The review identifies critical areas for future research, including the need for more sophisticated approaches to learning structured representations, integrating episodic memory mechanisms, and developing adaptive updating algorithms for working event models. By synthesizing insights from both theoretical frameworks and computational implementations, this review aims to advance our understanding of human event comprehension and guide future modeling efforts in cognitive science.

# 2. Introduction

## What is comprehension?

Since the advent of the concept "situational model" (Johnson-Laird, 1983), many researchers have argued that the construction of a coherent situation model is tantamount to the successful comprehension of a text (van Dijk and Kintsch 1983; Glenberg and Langston 1992; Graesser, Millis, and Zwaan 1997; Graesser, Singer, and Trabasso 1994; Zwaan, Graesser, and Magliano 1995). This change in the definition of the notion of comprehension shifts the research problem from the general, "How do readers comprehend a text?," to the more specific, "How do readers construct a coherent situation model?" (Zwaan and Radvansky 1998). This shift in focus underscores the importance of coherence in the mental representation of a text. Coherence itself can be further dissected into two: local and global coherence. Local coherence pertains to the immediate connections made between adjacent units of text, such as words, ideas, or sentences. These local connections facilitate a reader's understanding of the text in small chunks. Global coherence, on the other hand, refers to the organizational structure established within a broader discourse context, where local chunks of information are organized and interrelated into higher order chunks (Graesser et al. 1994). Both types of coherence emphasize the process of connecting disparate elements within a text to form a mental representation of interrelated elements.

Cognitive scientists have developed theories of the representation of knowledge, characterizing the way in which knowledge is structured so that progress can be made toward other important questions: how is old memory employed in the acquisition of new memory? How does structured knowledge facilitate comprehension? How does our current knowledge state modulate our actions? (Rumelhart and Ortony 1977). Accounts of knowledge representation proposed that knowledge is represented as frames

(Minsky 1974), scripts (Schank and Abelson 1977), schemata (Bobrow and Norman 1975; Rumelhart and Ortony 1977), or schemas (Graesser and Nakamura 1982). It has become a well-accepted view that the prior knowledge that readers possess plays a central role in reading comprehension. Theoretical accounts of discourse comprehension adopted these accounts of knowledge representation and started asking questions about processes supporting online comprehension and memory formation. Event comprehension theories moved beyond text to other modalities and incorporated prediction mechanisms, along with mechanisms to learn event schemas from repeated experiences.

The purpose of this review is to better understand the theoretical assumptions that have been made by event comprehension theories and to examine contemporary event comprehension models to identify necessary, unnecessary, and missing dimensions. I will also review discourse comprehension theories to identify relevant theoretical assumptions that can help development of event comprehension theories. There have been many discourse and event comprehension theories proposed over the last several decades--far too many to review here. Hence, to narrow the task, I focus here on discourse comprehension theories that are relevant to the development of event comprehension theories. Moreover, event comprehension models are selected such that they are distinctive enough from each other and provide good starting point for comparisons.

## 3. Conceptual models of discourse and event comprehension
### Classic discourse comprehension accounts
In the early days, researchers use discourses (texts) to study comprehension and memory processes. Part of the reason is that texts provide a convenient modality for presenting stimuli, and another reason is that text comprehension is supported by a variety of cognitive processes, many of which are assumed to be relevant to comprehension beyond text (McNamara and Magliano 2009:9). In what follows, I will briefly describe key features of four theories of discourse comprehension: Construction-Integration, Event Indexing Model, Causal Network Model, and Resonance model, and identify points that are relevant to modeling event comprehension. For a more comprehensive review of discourse comprehension theories, please see (McNamara and Magliano 2009).

The Construction-Integration (CI) model focused on the processes that give rise to discourse comprehension. The model essentially assumes a two-step cognitive process: construction and integration. Initially, when you read a sentence, the Construction phase kicks off by activating information in the text and related knowledge. Then, this cloud of activated concepts undergoes a refining process—Integration—where closely related concepts are strengthened, while isolated concepts are weakened. This leads to a stable state where the activation of concepts doesn't change much. The CI model assumes that a sentence that is read comprises three levels of representation: surface structure, propositional textbase, and situation model (van Dijk and Kintsch 1983; Kintsch 1988, 1998; Sanjose, Vidal-Abarca, and Padilla 2006). Of interest is the propositional textbase level, which is assumed to be the fundamental unit of representation. A proposition consists of a "predicate" and its "arguments," and typically conveys a single complete idea in the form predicate(argument, argument). For example, the sentence "a dog chases a cat" can be represented symbolically as chase(dog, cat). Discourse and event comprehension computational models must be able to address the challenge of extracting propositions from texts or videos. The CI model argues that 'argument overlap' is a sufficient basis for connecting propositions, effectively linking mental representations to achieve coherence (Giora 1985; Magliano, Trabasso, and Graesser 1999; Zwaan et al. 1995).

Event-indexing model (Zwaan and Radvansky 1998) offers an account of how individuals comprehend narratives, and the nature of situational models. One of its foundational assumptions is the cognitive system's heightened sensitivity to dynamic events—changes in states or conditions—rather than static elements. Event Indexing model explicitly proposes that humans monitor changes in five dimensions: time, space, causality, intentionality, and protagonist, which are dimensions prevalent in contemporary discourse comprehension theories. These dimensions are not exhaustive, but they provide a good set for event modelers interested in goal-directed activities. According to the Event-Indexing model, mental model construction involves the "current model" and the "integrated model". The "current model" is a mental representation of the focal event being described by the sentence that is actively being read. On the other hand, the "integrated model" serves as an evolving situation model that continuously gets updated as new "current models" are incorporated into it. The current model can be mapped onto the integrated model to the extent that readers perceive and infer how the two models are related in terms of time, space, causality, motivation, and protagonist. A theoretical distinction between the Event-Indexing and CI models is that linking propositions (i.e. mental representations of discourse constituents) is driven by event and causal relations primarily conveyed by verbs, not by the overlap of arguments.

Another discourse comprehension model that concerns the nature of the situational model is the Causal Network Model. The Causal Network Model aimed to understand the role of causality in the comprehension of text and discourse (Suh and Trabasso 1993; Trabasso and van den Broek 1985; Trabasso and Sperry 1985; Trabasso, Van Den Broek, and Suh 1989; Trabasso and Wiley 2005). Central to this model is the premise that comprehension primarily arises from causal reasoning--causal inferences are the main drivers for building a coherent situational model. The model lays out an analytic framework for researchers to identify causal relations in narratives, based on a general transition network. Propositions are first classified to categories: settings, events, goals, attempts, outcomes, and reactions. "Settings introduce characters, time, and place, while events are unintentional changes in state. Goals represent the desired states of characters, and attempts are actions taken to achieve these goals. Outcomes are the direct results of these intentional actions, and reactions encompass changes in the psychological states of characters. Under the causal network model, hierarchical representations of discourses can be viewed as an emergent property of causal inference processes, with goals motivating subgoals which motivates actions or enables outcome. It is important to note that not all texts afford a hierarchically structured representation and as such, this is driven by features of the discourse. Relatedly, Event Indexing Model and Causal Model assume that verbs carry more information, whereas CI model assumes that nouns carry more information and serves to connect propositions. This difference can be explained by types of discourse: CI's model focuses more on expository texts while the other models focus more on narrative texts. This highlight the an important point that is relevant to modeling event comprehension: discourse or event features dictate processing assumptions (inductive biases) in theories and computational models.

The Resonance model was proposed to explain how information that is relatively far from the focal sentence (e.g. in the last paragraph) in a text can be reactivated, even when it is no longer present in working memory (Cook and O'Brien 2014; Myers et al. 1994; Myers and O'Brien 1998; O'Brien and Cook 2016). It operates on the principle that long-term memory traces resonate depending on their match to the content of working memory. This matching is influenced by the overlap in semantic and contextual features among concepts and the argument overlap of propositions. In this model, signals from working memory are perpetually being sent to long-term memory and the degree of activations of long-term memory depends on featural overlap with working memory. It emphasizes that the resonance process is not an isolated event but is continual. Contrary to the Resonance model, the minimalist hypothesis (McKoon and Ratcliff 1992) contends that inferences to distant text are made only when there's a break in local coherence.  Local coherence breaks can occur when propositions of the current sentence fail to share arguments with propositions from immediately preceding sentences (van Dijk and Kintsch 1983), or when a causal relation is absent between events described in the current sentence and events in the immediately preceding sentences (Van Den Broek 1990). One important question in modeling event comprehension is how episodic memory and working memory interact so that stored episodic memories can facilitate comprehension. The two views presented here provide potential mechanisms.

This section discusses four theories in discourse comprehension: the Construction-Integration (CI) model, the Event-Indexing Model, the Causal Network Model, and the Resonance model. These theories highlight key questions in discourse comprehension research that are relevant to event comprehension theories: How are causal relations and the components of activities learned? How can structured representations be constructed from texts or videos? How are event dynamics learned? And, how do working and episodic memory interact to facilitate comprehension?

## Contemporary event comprehension accounts

Researchers started to develop event comprehension theories applicable to other modalities (audio/visual). Although discourse models assume predictive inferences are strategic processes and readers do not always engage in predicting the future (Allbritton 2004; McDaniel, Schmalhofer, and Keefe 2001), event comprehension accounts assume that prediction is a key component in comprehension and memory processes and people generate predictions continually during comprehension. In what follows, I will describe prevalent event comprehension theories while identifying key computational challenges that event modelers need to address. Then, I will try to set up a framework to describe contemporary computational models of event comprehension.

Event Segmentation Theory (EST) (Zacks et al. 2007) posits that our cognition is inherently predictive. It suggests that humans and animals are continuously trying to forecast what will happen in their immediate future. This is useful because it helps us prepare for what's coming. In real life, it is common to encounter situations where all necessary information is not available. This could be due to lapses in attention, inability to see certain things, or even the inherent limitations of our sensory systems. To bridge these gaps, the mind employs "event models." These are mental representations that blend immediate sensory input with long-term schematic knowledge (event schemas)**,** offering a stable picture of the immediate situation. Event schemas encode knowledge about how an event (e.g. going to a restaurant) typically unfold, and they are acquired by learning from repeated experiences. This raises an important question: how are event schemas represented and learned with experiences? Event models encode what is currently relevant to one's goals and are multimodal, meaning they combine different kinds of sensory. One of the key features of event models is that they are generally kept stable by isolating them from continuous sensory input. This is crucial because it allows the model to provide a consistent interpretation of the world, especially when faced with transitory information. However, this doesn't mean that working models are static. They need to be adaptable to changing circumstances. When and how are the stable working event models updated? EST suggests that the mechanism for this adaptability lies in monitoring "prediction error," the difference between our anticipations and actual occurrences. When this error spikes, it indicates that our current working model is no longer sufficiently accurate, triggering an update. This updating process opens the event model to new information for a brief window, both from our current sensory perceptions and from our long-term understanding of similar situations (schematic knowledge), thereby forming a revised mental representation. Updating event models based on prediction errors implements a form of cognitive control: when to actively takes in sensory input and update working memory content. Event Segmentation Theory (EST) proposes that individuals maintain event models across different timescales simultaneously. This raises a mechanistic question: How does coarser event models influence finer event models, and vice versa? In other words, how does information flow across hierarchies?

Event Horizon Model (Radvansky and Zacks 2014) elaborates EST further. According to the Event Horizon Model, humans innately track the cause-and-effect relationships between events. This aids in forecasting future events and acting adaptively. For instance, if you learn that a specific cloud formation often leads to rain, you might carry an umbrella the next time you see those clouds. Causality can also help segment events. From the viewpoint of EST, causal breaks serve as cues for event segmentation because they correspond to moments where prediction becomes difficult. Additionally, when there is a causal discontinuity in a narrative text, a plausible cause must be inferred, and this causes a new event model to be created**.** Causal relations are part of schematic knowledge, and the centrality of causality in discourse and event comprehension theories raise an important question: how causal relations are learned?

Kuperberg (2021) proposed a hierarchical generative framework to model event comprehension. The core of this proposal is a three-level hierarchical generative model. Level 3 represents probabilities over possible goals of the agent. Each goal is associated with a relevant schema, and a range of possible end states. Event models are represented at level 2, encoding what has happened, what is happening, and what might happen next. The predicted and observed stimuli are represented at level 1. This framework provides an account how information flows across different levels of representations. Within each two-level, information represented at the higher level is actively propagated down, through feedback connections, to reconstruct activity at the lower level: goals probabilistically predict possible event models, event models probabilistically predict possible upcoming stimuli. The reconstructed state of activity at the lower level (e.g. level 1) is then subtracted from the state that is induced when new stimuli appear. Only the difference in activity (i.e. errors) is passed back up to the higher level (e.g. level 2) via feedforward connections (Clark 2013; Friston 2005; Friston et al. 2016; Rao and Ballard 1999). When this bottom-up prediction error reaches the higher level (e.g. level 2), it induces an update in level 2 representations. As each new input becomes available from the environment, this process is repeated until the magnitude of the bottom-up prediction error is minimized. For example, when a new scene is observed at the first level, prediction error is computed and fed to the second level, and the event model is updated. The newly inferred event model, in turn, probabilistically predicts upcoming scenes at that moment in time. In short, higher levels project predictions to the lower levels, and only errors are propagated backward. Under this framework, an event boundary is experienced by the observer when there is a shift the probability distribution over possible goals—when the observer believes that the goal of the agent has changed. This detection can be achieved through two means. First, the model monitors prediction errors and update goal distribution because of errors propagated from level 2. Second, the model monitors uncertainty over goals, which rises when end state associated with the current goal is attained, to update goal distribution. This raises a question: how are end states associated with a goal or an event schema is learned?

A goal is a state of affairs that an entity acts to bring about, from an initial state of affairs. Goal is important because it causes the observed sequence of events, and that "to truly comprehend an action sequence (infer its underlying latent cause and explain the input), we must be able to infer the agent's goal". The notion of causality is consistent with Trabasso and colleagues who proposed that comprehension is driven mainly by causal reasoning. Recall that goal motivates other goals, attempts, and outcomes, and this is only one type of causal relationship--motivational. Other types of causal relationships are enabling, physical, and psychological. Thus, to explain input by inferring its underlying latent cause, one must not only infer the agent's goal but also the settings that enable attempts, and the inferred circumstances, which all can be cast as latent causes producing observed actions. So, it might be instructive to reconceptualize level 3 in the hierarchical generative model as not just representing goals but also settings and inferred contextual information or other causes of the observed sequence of events. Furthermore, we use terms like goals, settings, and circumstances to pump the intuition. However, as we have learned from discourse comprehension, the content of latent causes should depend on the type of discourses or activities. This reconceptualization doesn't assume the exact content each level represents but let the model learn relevant dimensions at each level, which is typical of connectionist approach to cognitive modeling (McClelland et al. 2010).

In a hierarchical model, high-level representations can be built in two primary ways using bottom-up information propagation. First, features become increasingly abstract as they move up the hierarchy, as seen in convolutional or recurrent neural networks. In this approach, higher-level units process and compress information from lower levels (this view is adopted by early vision scientists studying the visual cortex, but see (Rao and Ballard 1999). Second, higher-level representations are created by inferring what best explains the lower-level data based on previously learned knowledge. This method uses a generative model to reconstruct and compare lower-level representations, adjusting higher-level ones based on any prediction errors. So, while feedforward connections in the first method carry lower-level representations, in the second they carry prediction errors, which is the assumption in predictive coding frameworks (Clark 2013; Friston 2005; Kuperberg 2021). This poses two important questions for modeling event comprehension: Whether both representations and errors are propagated to the next level, and what their roles are in updating representations of the higher level.

*Theories of discourse and event comprehension pose mechanistic questions that computational models of event comprehension should address. The questions are:*

1. How are event schemas represented and learned? This includes:
    a. How are events represented hierarchically?
    b. In a hierarchy, how does information flow across levels and within a level? How are those dynamics learned?
    c. How are end states learned?
2. When and how are working event models updated?
3. How are components of an activity learned?
4. How does episodic memory support online comprehension?
5. How selective attention controlled?
6. How are structured representations constructed?

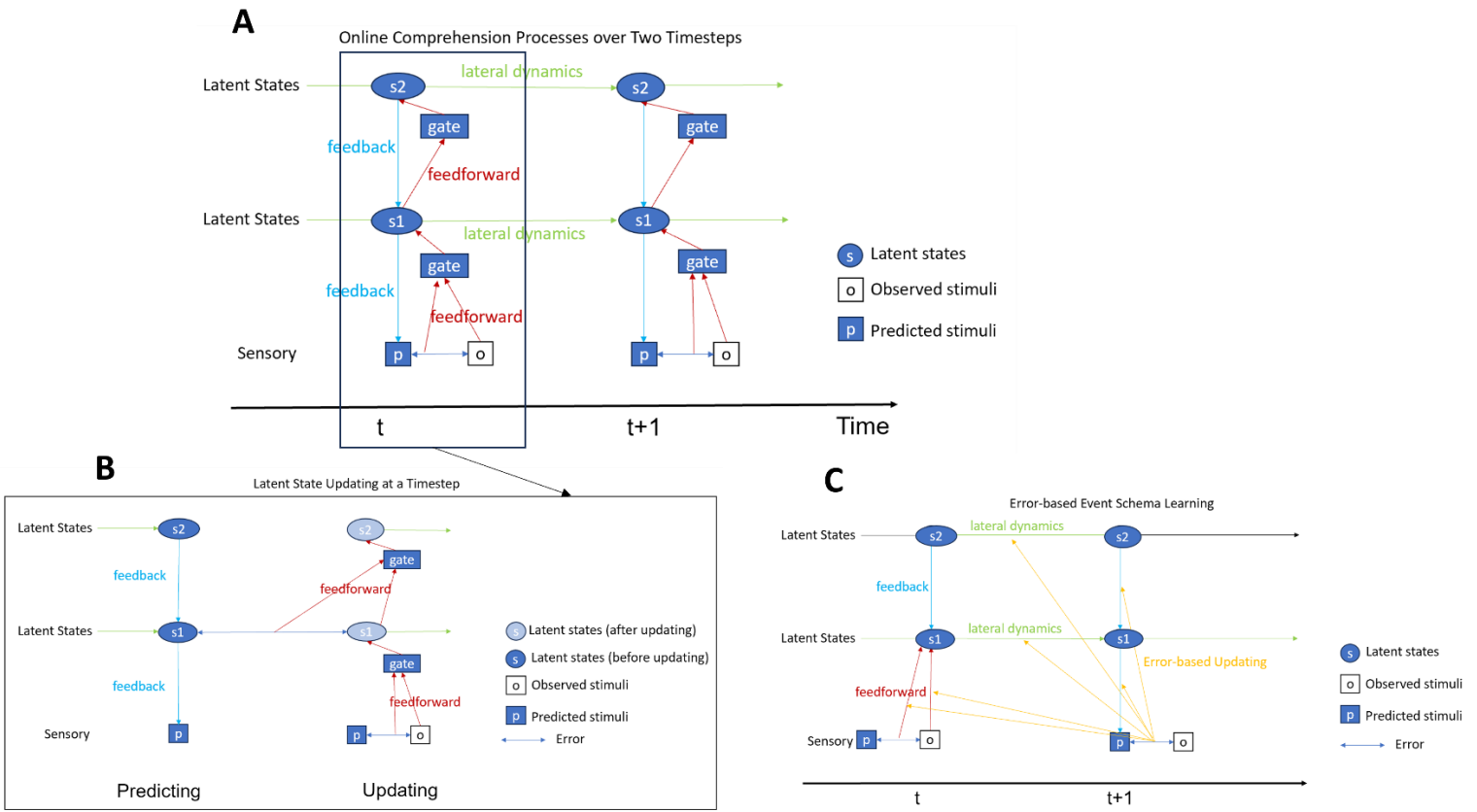

Figure 1: A place holder framework illustrating different comprehension processes. A) Online comprehension processes over two timesteps: Top-down predictions, lateral dynamics, and gated feedforward propagations in a three-level model. Theories differ in the nature of these dynamics. Solid shapes are unobservable, empty shapes are observable. B) The two-step process: predicting of sensory information and updating of latent states. Predicting step: prior latent states influences current latent states (lateral dynamics arrows), which generate predictions (feedback arrows). Updating step: latent states are updated based on new sensory observations and errors (feedforward arrows). C) Learning of event schemas: The learning involves making future predictions and update feedback, feedforward, and lateral relations based on prediction errors. The targeted arrows (blue, red, green) signify potential connections subject to error-based learning.

To evaluate how different computational models of event comprehension address mechanistic questions posed by theories, I use a placeholder model in Fig. 1. This placeholder model has arrows indicating interactions among representations across levels. Importantly, the nature of these interactions will be different for different computational models, having a placeholder model helps comparing models more easily. Fig 1A illustrates the online comprehension processes over two timesteps in a hierarchical model with three levels. Interactions between levels are defined by top-down feedback, lateral dynamics within each level, and feedforward propagations (including both representations and prediction errors). Feedforward propagations could be gated before reaching a higher level so that representations on a higher level are more stable and operate in a longer timescale.

The nature of the gate is different among cognitive theories. For example, EST proposed that the gate is binarized, opening when prediction errors are high and remaining open for a brief period at the start of the event, thus implementing a step gating function. Under predictive coding with Bayesian updating (Friston et al. 2016; Kuperberg 2021), the gate is controlled by the uncertainty over latent states, thus implementing a smooth gating function. The gate function can also be learnable (Gumbsch, Butz, and Martius 2021; Lu, Hasson, and Norman 2022), opening when it's beneficial for predicting the future. Fig. 1B illustrates predicting and updating operations within a timestep. In Predicting step, before new sensory input comes in, current latent states are determined by latent states in the previous timestep (green arrows) and top-down feedback from higher levels (cyan arrows); current latent states then generate sensory predictions (cyan arrows). In Updating step, as new observations come in, sensory predictions are compared with observations and prediction errors are computed. Errors and representations of sensory observations (e.g. representations in the retina) are propagated upward to update representations at the next level (feedforward--red arrows). The degree of updating at this level is conceived as prediction errors and, along with the updated representations, they get propagated to the next level. Fig. 1C illustrates how event dynamics (top-down, bottom-up, and lateral dynamics) are learned. Errors are propagated to update feedback, feedforward, and lateral connections (yellow arrows). The nature of error-based updating is different across theories and computational models. Here, the set of target arrows indicate potential connections that could be updated. Episodic memory and attentional control are not present in the current placeholder framework for sake of tractability, I will discuss these two processes in the next sections.

## 4. Computational models of event comprehension

In this section, I will describe how each model addresses questions posed in the previous section through operationalizing placeholders in Fig. 1. Note that all computational event models are tasked to predict the next sensory information given what has been observed.

### REPRISE

REPRISE (Butz et al. 2019), retrospective and prospective inference scheme, is introduced as a model that learns temporal event-predictive models of dynamical systems by inferring unobservable contextual event states. For the purpose of this review, it can be thought of as a model passively viewing a sequence of actions performed on the object and object locations, while object identity switches occasionally and unbeknown to the model. At a given timestep, the model is tasked with predicting the location of the object, given the object's current location and an action input. There are three types of objects, and depending on the type of object the action input might have different effects on the object's location. To learn to predict well in this setting, the model needs to infer which object it is viewing. A priori, the model does not know how each object reacts to action input—the dynamics associated with each object. Thus, the model needs to learn these dynamics. This setting is similar to where a human baby observes another human performing an activity: the baby needs to learn the dynamics associated with each type of event (i.e. learning event schemas), and also needs to infer which event schema is appropriate at the moment to make good predictions about the agent the baby is observing.

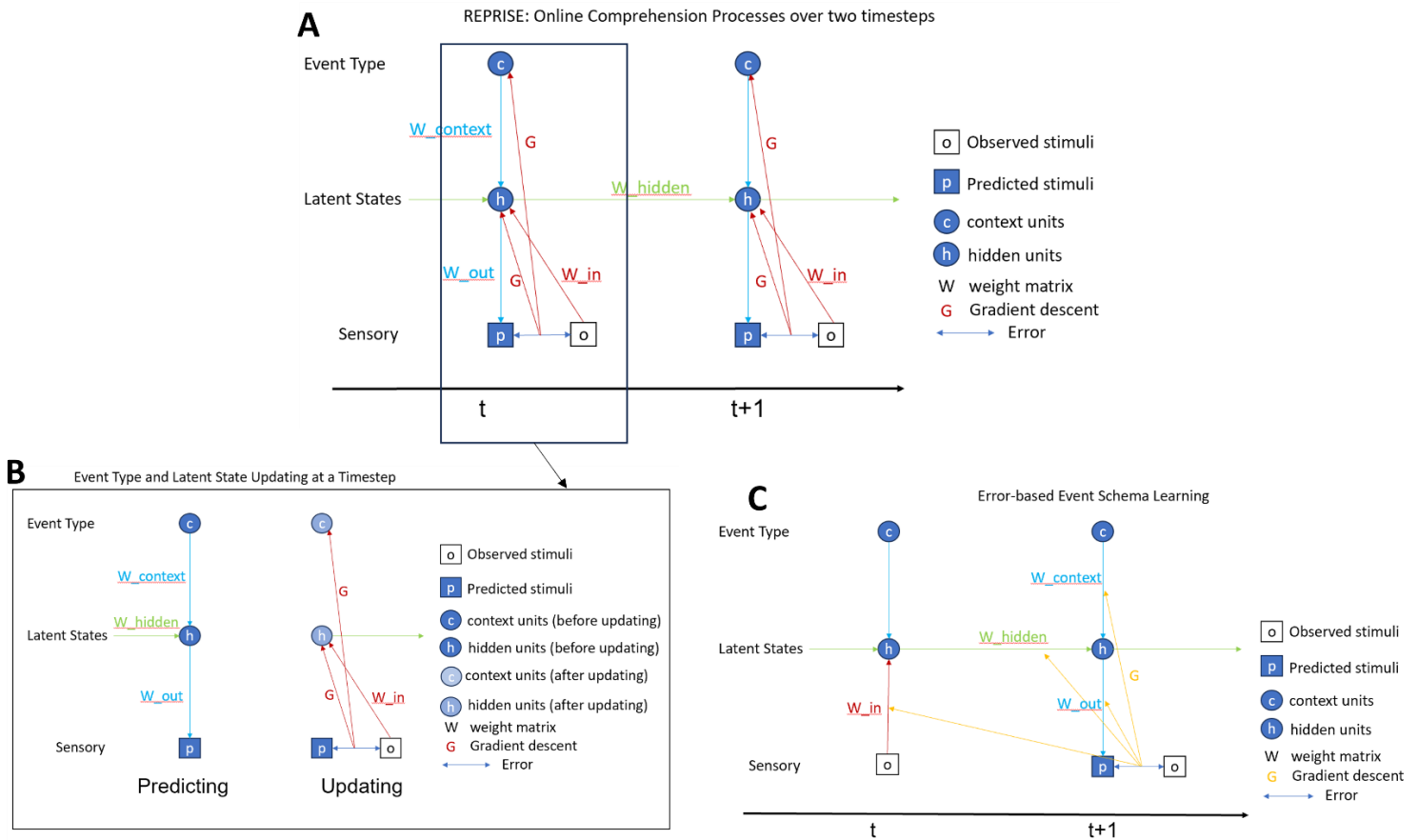

Figure 2: Mapping three-level REPRISE model to the placeholder framework. A) Online comprehension processes over two timesteps. Feedback are represented by two matrices (W_context and W_out), lateral dynamics are represented by W_hidden, feedforward is represented by weight matrix W_in and gradient vectors G. B) The two-step process: prediction of sensory information and updating of context vector and hidden layer vector. In Predicting step, top-down feedback is projected from context layer vector (level 3) to hidden layer vector (level 2), and activations of the hidden layer vector in the previous timestep also influence activations of the hidden layer vector in the current timestep (lateral dynamics). Hidden layer vector (level 2) generates a sensory prediction at level 1. Thus predictions of the sensory stimuli are influenced by both latent states (hidden layer vectors) and contexts (context vector). In Updating step, sensory prediction errors and sensory observations are both propagated upward to level 2. Sensory prediction errors (level 1) are propagated to update context layer vector (level 3). Prediction errors guide the adjustments to activations of hidden layer vector (world states) and context vector (event types). Notably, updates to the hidden and context vectors are continuous and not gated, with distinct propagation patterns for errors and representations for hidden layer vector (both sensory prediction errors and sensory observations) and context layer vector (only sensory prediction errors). C) Learning of event schemas: Observations and predictions are juxtaposed, with ensuing errors directing modifications to the RNN's weights (W_out, W_in, W_hidden) and context's weights (W_context). Note that gradient descents in the event type and latent state updating step in 2B (red arrows) are taken over context vector and hidden layer vector, which are different from gradient descents in the event schema learning 2C (orange arrows) which are taken over weight matrices.

*How are event schemas presented and learned?*

**Hierarchical structure and information flow (top-down, bottom-up, and lateral dynamics):** In REPRISE, there are three levels of representations: sensory information (location) is represented at level 1, latent states of the object (e.g., velocity, acceleration, etc.) are represented at level 2, and event type (object type) is represented at level 3 (Fig. 2A). Context vector in level 3 modulates representations at level 2 (matrix W_context), which in turn generate top-down predictions at level 1 (matrix W_out) (Fig. 2B, Predicting step). Level 1 to level 2 feedforward contain both errors (gradient descent vector G) and representations (matrix W_in), which are used to update latent states (Fig. 2B, Updating step). Notably, level 2 does not propagate information to level 3. Instead, level 1 sensory errors are propagated to level 3 (gradient descent vector G) to infer the current object being handled. Lateral dynamics define how representations of the previous timestep influence representations of the current timestep within a level. The W_hidden weight matrix defines how latent states evolve—level 2 lateral dynamics. Level 3 representations are copied to the next timestep, while level 1 representations in the previous timestep don't influence level 1 representations in the current timestep.

**Schema learning:** In REPRISE, event schemas are defined as four weight matrices connecting levels (fig. 2C): The RNN's weights (W_out, W_in, W_hidden), and the projections from the context vector to hidden layer representations (W_context). Even though all schemas share the same four weight matrices, they differ in their activations in the context vector. Different activations in the context vector influence hidden layer vector differently, which in turns makes different predictions about the upcoming sensory information given the same action input. REPRISE predicts current object's location given current beliefs about event types (represented in the context vector) and states of the object (represented in the hidden layer vector). Observations are compared to predictions, and errors are propagated back to adjust RNN's weights (W_out, W_in, W_hidden) and the context's weights (W_context) (fig. 2C). Intuitively, adjusting the four weight matrices can be conceptualized as learning to predict what should be happening given the current beliefs about states of the world (hidden layer vector) constrained by current beliefs about the event type (context vector).

*When and how working event models are updated?*

In REPRISE, working event models are represented by the context vector and the hidden layer vector. During comprehension, prediction errors, computed based on predicted and observed locations, are propagated to adjust activations of the hidden layer vector (representing states of the world) and the context vector (representing event types) (Fig. 2B, Updating step). Note that hidden layer vector and context vector updating are continual and not gated in REPRISE. Implicit in the model, event boundary can be defined as a sharp change in the patterns of the context vector representing beliefs about event type.

## SEM

Structured Event Memory (SEM) model (Franklin et al. 2020) was introduced to model human event segmentation, memory, and generalization. For the purpose of comparisons with other models on key comprehension processes, I skip memory encoding and retrieval processes here and will revisit them in the next section. The model passively views a sequence of events, each event containing a unique sequence of scenes (frames). At a given timestep, the model is tasked with predicting the next scene in a naturalistic everyday event, given observed scenes. In the dataset, there are many types of events. To learn to predict well in this setting, the model needs to infer which event the agent is performing (e.g. making a bagel) and it also needs to learn how the event typically unfolds—learning the dynamics of each event type. A priori, the model does not know the dynamics associated with each event type. Thus, the model needs to learn these dynamics.

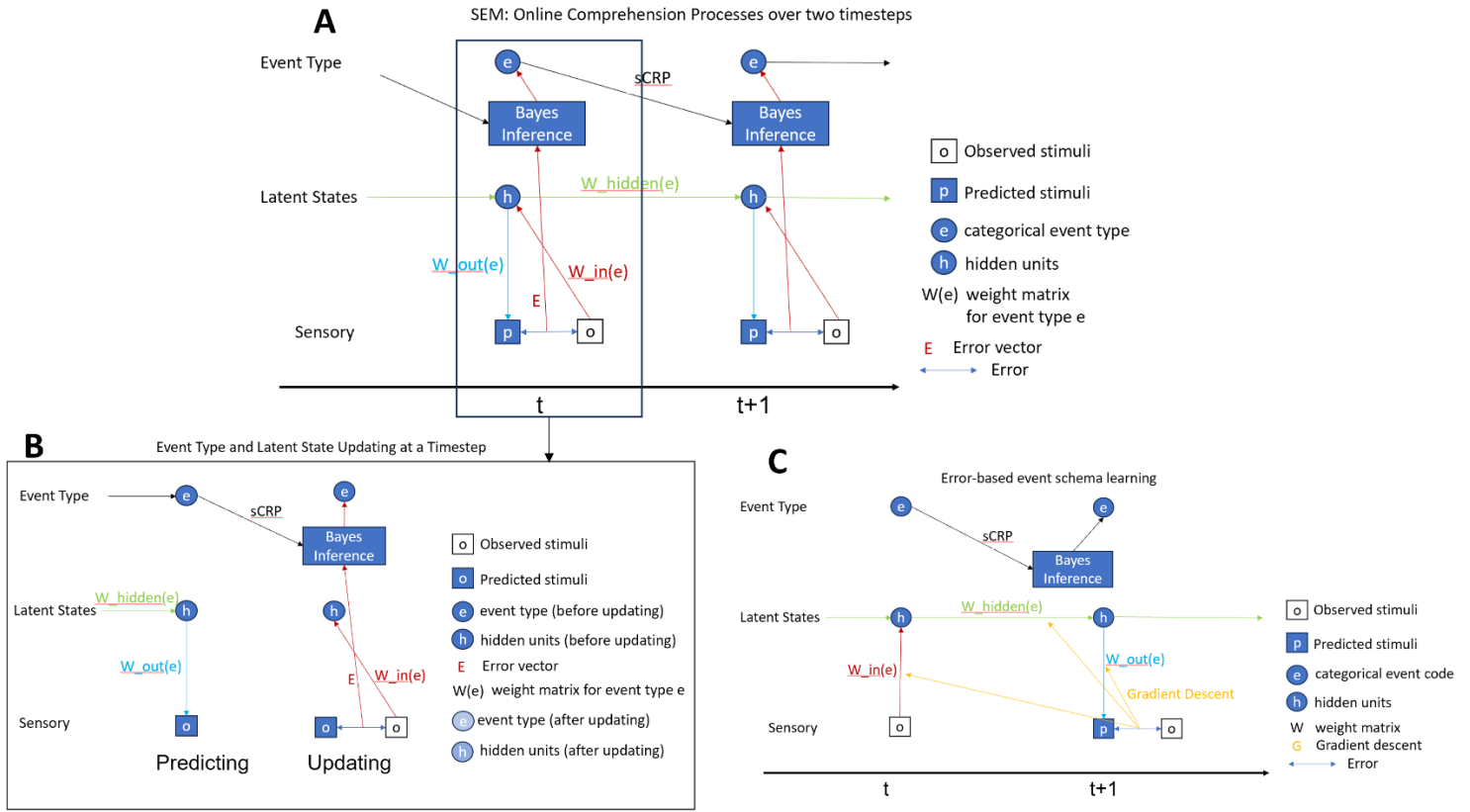

Figure 3: Mapping the three-level Structured Event Memory (SEM) model to the placeholder framework. A) Online comprehension processes over two timesteps. Feedback are represented by matrix W_out(e), lateral dynamics are represented by matrix W_hidden(e), feedforward is represented by weight matrix W_in(e) and error vector E. Event schemas in SEM are represented by unique RNNs for each event type, differing from REPRISE representing event types by a context vector. Each event type has its own weight matrices: W_in(e), W_out(e), and W_hidden(e). B) The two-step process: predicting of sensory information and updating of latent states and event types. Prediction step: prior latent states influence current latent states (lateral dynamics arrow in green), which generate sensory predictions (feedback arrow in cyan). Updating step: latent states are updated based only on sensory observations (feedforward arrow in red), and event types are updated based on sensory prediction errors (feedforward arrow in red). SEM infers event type through approximate Bayesian inference. Event types are evaluated based on a combination of the sticky Chinese Restaurant Process (sCRP) prior that accounts for temporal contiguity within events and the likelihood inversely proportional to prediction errors. C) Learning of event schemas: The learning involves making future predictions and adjusting the RNN weights according to prediction errors. Notable, only the weights of the current event type are learned.

*How are event schemas represented and learned?*

**Hierarchical structure and information flow (top-down, bottom-up, and lateral dynamics):** In SEM, there are three levels of representations: sensory information is represented at level 1, latent states of the current event are represented at level 2, event type is represented at level 3 (Fig. 3A). Hidden layer vector (level 2) generates top-down predictions at level 1 (Fig. 3B, Predicting step). Level 1 to level 2 bottom-up propagations contain only representations (W_in), which are used to update latent states of the current event (Fig. 3B, Updating step). Level 2 does not propagate information to level 3. Instead, level 1 errors are propagated to level 3 to infer the current event type the agent is engaging with. The W_hidden weight matrix defines how representations of the current states of the world (latent states) evolve—level 2 lateral dynamics. Level 3 event type in the previous timestep influences level 3 event type in the current timestep through the sticky Chinese Restaurant Process (sCRP), which is used as a prior over event types. Level 1 representations in the previous timestep don't influence level 1 representations in the current timestep.

**Event schema learning**: In SEM, event schemas are represented by RNNs (W_in, W_hidden, W_out), one RNN for each schema (Fig. 3C). SEM learns a unique RNN for each event type (W_in(e), W_out(e), W_hidden(e)). At any timestep, SEM makes predictions about upcoming sensory information. Prediction errors are used to updated RNNs' weight matrices, effectively learning event schemas. Notably, at any updating step, only the active RNN representing SEM's belief about the current event type is updated.

*When and how working event models are updated?*

In SEM, working event models are represented by the hidden layer vector and the event type. Hidden layer vector is updated gradually when event type of two consecutive timesteps are the same (fig. 3B, Updating step), but it is updated dramatically when two consecutive timesteps have different event types. Event type is inferred by approximate Bayesian inference (fig. 3B, Updating step) process. Prior over event types (level 3) are defined as a sticky Chinese Restaurant Process (sCRP), which assumes temporal contiguity within events. Temporal contiguity is the idea that two scenes that occur close together in time tend to belong to the same event. Likelihood over event types is inversely proportional to prediction errors: event types that make predictions close to observations at a given time are given high likelihood. Event type that has the highest posterior probability is selected for the current timestep. In SEM event boundary is conceptualized explicitly as the transition between event types.

## A model of when to retrieve memory

The Lu model was introduced to model human episodic memory control. An event is a sequence of interleaved features and queries. For example: day=weekday, barista=?, sleepiness=sleepy, mood=?, weather=sunny, coffee=?. At each timestep, the model observes one feature and is asked to answer a query (e.g. barista=Bob or Eve?). Critically, answers to queries are determined by features. For example, if the day feature is weekday, barista is Bob (this schematic knowledge must be learned by the model). At the end of a sequence of features (end of an event), the model's working memory is copied to episodic memory, and its working memory is flushed before encountering another event. Thus, at a given timestep, the model can answer a query (e.g. whether the barista is Bob or Eve) based on features that it has observed within the current event and maintained in its working memory (e.g. it is a weekday) or based on features that it has observed in another events and stored in its episodic memory. This setting is similar to a naturalistic setting where two friends go to a Wendy and one of them predicts what the other will order when they are at the counter, prediction might rely on their recent conversation when they approach the counter and the other one said they are in the mood for chicken, or prediction might rely on their conversation yesterday and the other one said they are vegetarian. Because episodic memory might contain lures, and retrieving requires cognitive resource, it's not optimal to retrieve stored memory traces all the time. The model needs to learn a policy when to retrieve episodic memory.

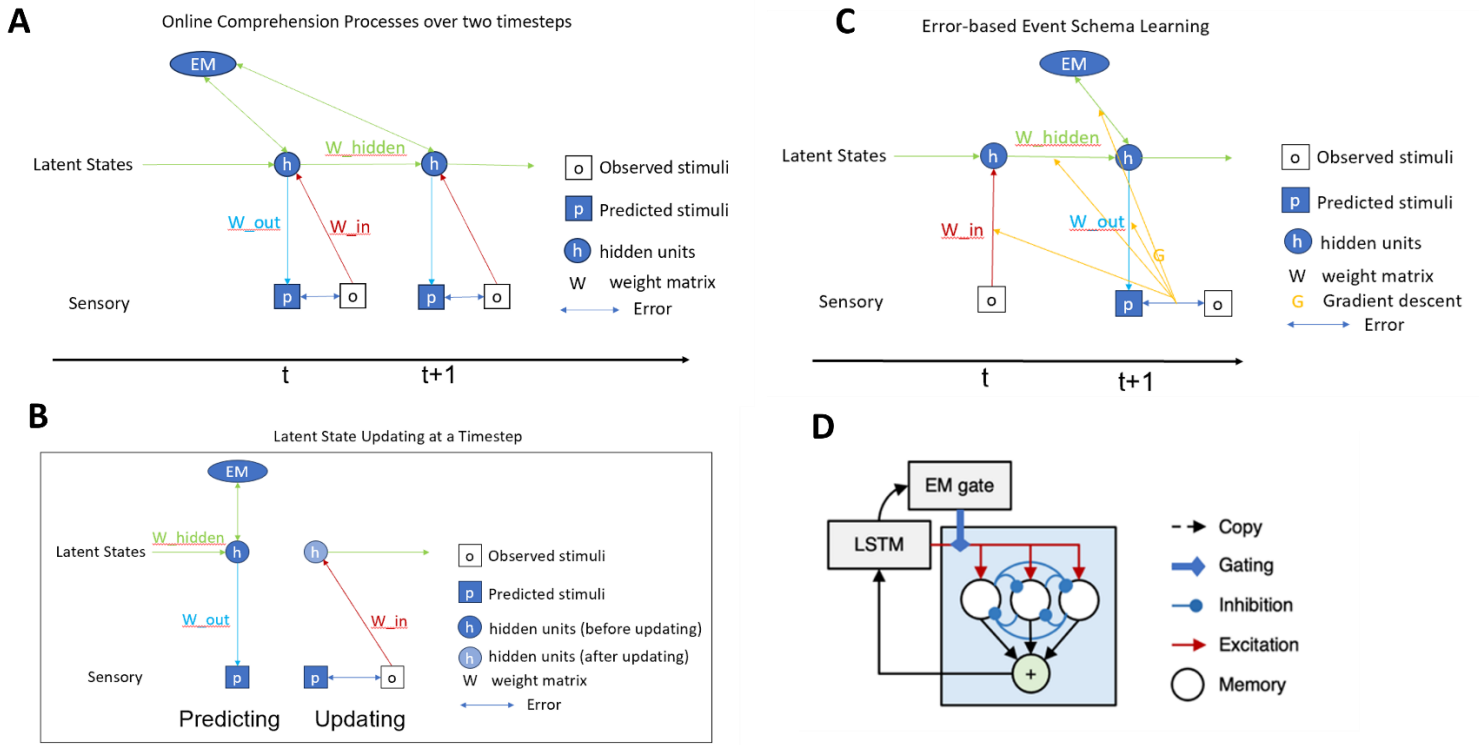

Figure 4: Mapping the two-level Lu model to the placeholder framework. A) Online comprehension processes over two timesteps. Feedback is represented by matrix W_out, lateral dynamics are represented by matrix W_hidden, feedforward are represented by matrix W_in. Interaction between hidden states and episodic memory is represented by the double headed arrow and elaborated in D. B) The two-step process: predicting sensory information and updating hidden layer vector. In Predicting step, activations of the hidden layer in the previous timestep affect hidden layer activations (level 2) of the current timestep (lateral dynamics in green arrow). Depending on the EM gate, episodic memory could also affect hidden layer activations of the current timestep. Hidden layer vector then generates answers to the queries of the previous timestep (level 1). In Updating step, only sensory observations (not errors) are propagated upward to influence hidden layer vector (level 2). C) Learning of event schemas: The learning involves making future predictions and adjusting the W_in, W_out, and W_hidden weight matrices according to prediction errors. Notably, memory retrieval policy is also learned by the model. D) Working memory and episodic memory interaction: Episodic memory retrieval depends both on similarity between memory traces and current representations in LSTM (red arrows) and on the degree of uncertainty in LSTM representations (blue gate).

*How are event schemas represented and learned?*
**Hierarchical structure and information flow (top-down, bottom-up, and lateral dynamics)**: In the Lu model, there are two levels of representations: features and query answers are represented at level 1.

Latent states of the current event (what features have been observed and what the current query is), or working memory, are represented at level 2. Episodic memory can be conceptualized as an extension of level-2 representations: Whereas hidden layer vector is conceptualized as working memory (or working event model), stored vectors of past activations are conceptualized as episodic memory traces (fig. 4A, h represents working memory and EM represents episodic memory). At level 2, the content of working memory continually sends signals to episodic memory, which in turns sends back stored memory traces that are similar to the content of working memory (fig. 4B, Predicting step). Working memory representations generate top-down sensory predictions at level 1 (answers to queries).  Level 1 to level 2 propagations only contain representations (W_in) to update latent states in working memory (fig. 4B, Updating step). The W_hidden weight matrix defines how the content of working memory in the previous timestep influence the content of working memory in the current timestep—level 2 lateral dynamics. Like other RNN variants, in this model level 1 representations in the previous timestep don't influence level 1 representations in the current timestep.

**Event schema learning:** In the Lu model, there is only one event schema, which is represented by the RNN. RNN weight matrices this case encodes the feature-query pairs (e.g. if day=weekday, barista=bob). At a given timestep, the model makes predictions about upcoming sensory information (answer to a query) and use prediction errors to adjust its weight matrices (fig. 4C).

*When and how are working event models updated?*
Because the authors are primarily interested in how episodic memory contributes to prediction, the model learned one event schema instead of multiple event schemas at the same time. Hidden layer vector, which represents the content of working event model, is updated continually due to observed sensory information (fig. 4B, Updating step). Notably, this architecture allows stored episodic memories to modulate activations of the hidden layer vector, effectively operationalizing memory retrieval (fig. 4B, Predicting step).

*How does episodic memory support online comprehension?*
Episodic memory encoding is conceptualized as taking a snapshot of the current states of the world--copying activations of the hidden layer vector to a memory storage. When episodic memory is encoded is hand-coded by the authors. Episodic memory retrieval is implemented using a leaky competing accumulator (LCA) model (Lo and Ip 2021; Usher and McClelland 2001) (fig. 4D), which has been used in other memory models (e.g., Polyn, Norman, and Kahana 2009). Each memory trace receives excitation proportional to its similarity to the current activations of hidden units, and memory traces compete via lateral inhibition (fig. 4D)**.** Integration of memory traces to the hidden units (working memory) is controlled by a gate (blue gate in fig. 4D). Importantly, the policy when to integrate episodic memories is learned by the network. The network learned to retrieve episodic memories when there're gaps in its understanding (when it cannot predict) instead of retrieving from episodic memories all the time, with the potential cost of retrieving lures**.** In the model, the coherence of working memory is evaluated all the time, and this is operationalized by having hidden units sending uncertainty (incoherence) signals to the Episodic Memory module at all timesteps (black arrow from LSTM to EM in fig. 4D). The degree that episodic memories are integrated to working memory (event model) depends on the degree of coherence break in working memory (blue gate in fig. 4D). In summary, episodic memory integration into working memory depends both on similarity between memory traces and current representations in working memory and on the degree of coherence break in working memory. This view is consistent with both resonance hypothesis, arguing that memories continually resonate and integrate, and the minimalist hypothesis, arguing that integration are made by readers only when there's a coherent break in working memory.

## A sparsely changing latent state model that learns end states of events

The Gumbsch model was introduced as a model that learns when to update stable representations in working memory and where to direct attention. The Gumbsch model views sequences of robotic arm and object locations. Each sequence includes an ordered set of events such as "reaching the object", "grasping the object", and "transporting the object", where each event includes a sequence of robotic arm and object locations with unique trajectories. At a given timestep, the model is tasked with predicting the locations of objects and the robotic arm in the next timestep, given previous locations. Importantly, it does so by learning stable latent states (e.g. velocity of the robotic arm) and use them to makes predictions. To learn to predict well in this setting, the model needs to infer which high-level event it is viewing. A priori, the model does not know the dynamics associated with each high-level event. Thus, the model also needs to learn these dynamics. Again, this setting is similar to where a human baby observes another human performing an activity: the baby needs to learn the dynamics associated with each type of event (i.e. learning event schemas), and also needs to infer which event schema is appropriate at the moment to make good predictions about the agent the baby is observing. Moreover, the Gumbsch model also learns to predict situations in which latent states tend to change, and allocate attention appropriately based on uncertainties associated with these situations.

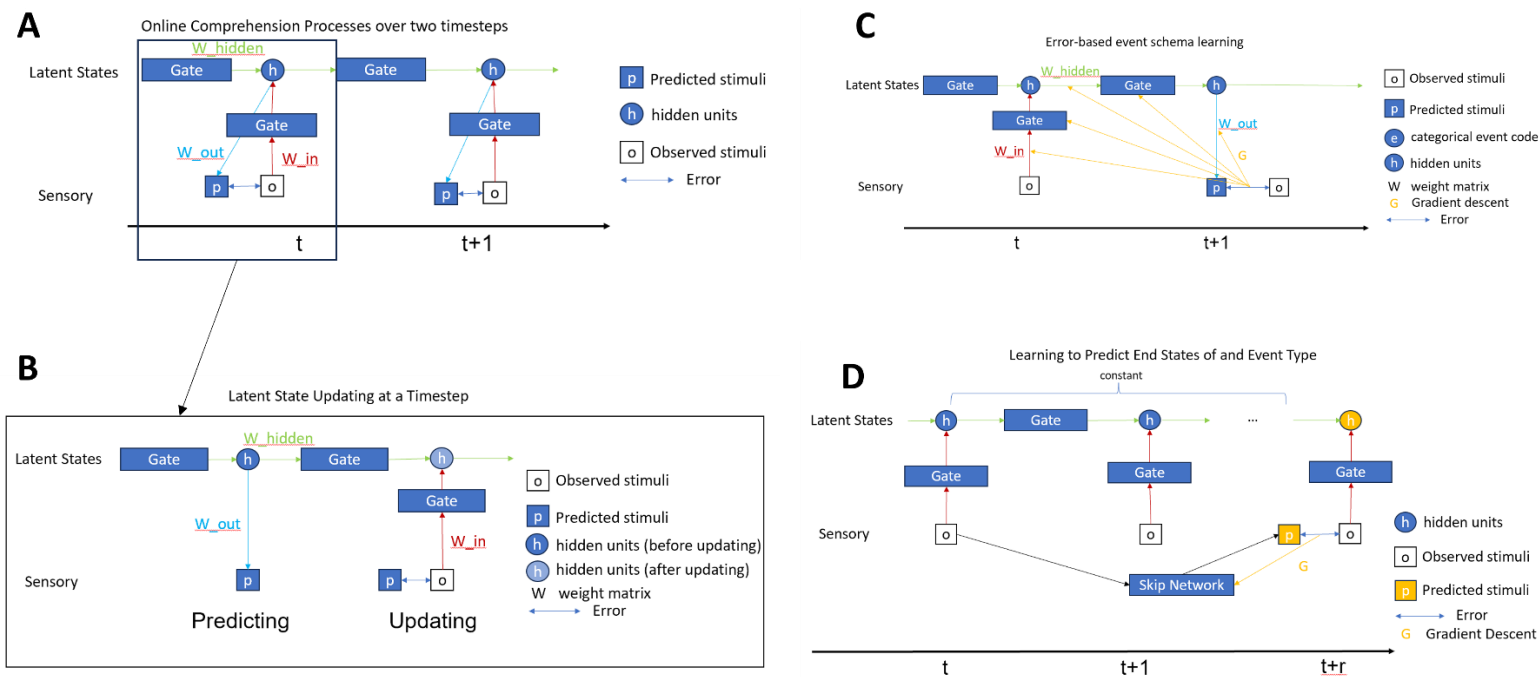

Figure 5: Mapping two-level event comprehension model proposed by Gumbsch et al. to the placeholder framework. For the purpose of this review, components related to actions, such as the inverse network, are omitted in this mapping. A) Online comprehension processes over two timesteps. Feedback is represented by matrix W_out, lateral dynamics are represented by matrix W_hidden, feedforward is represented by matrix W_in. Notably, feedforward and lateral dynamics are gated by a module that incentive changing latent states sparsely. B) The two-step process: predicting sensory information and updating hidden layer vector. In Predicting step, depending on the gate, activations of the hidden layer in the previous timestep could affect hidden layer (level 2) of the current timestep, which generates sensory prediction (level 1). In Updating step, only sensory observations (not errors) are propagated upward to influence hidden layer vector (level 2). This feedforward is gated. Because both lateral dynamics and feedforward are gated, hidden layer vector remains constant most of the time (not updated continuously like SEM or REPRISE). Even though the color of hidden units after updating is different from the color of hidden units before updating in the depiction, that does not necessarily mean that the values of two hidden layer vectors are different. The purpose is for this depiction compatible with other depictions. C) Learning of event schemas: The learning involves making future predictions and adjusting the weight matrices according to prediction errors. The gates are trained by imposing a cost of updating the hidden unit in the loss function, in addition to mean square of sensory prediction errors. This can be interpreted as the model is encouraged to update its latent state only when the benefit for prediction accuracy overweighs the cost of updating. These moments are signaled by patterns of activity in the hidden layer and the observed sensory input. D) Learning to predict end states. Sensory observations at timepoints where hidden layer vector changes (blue to yellow) are used as targets representing end states of and event type. A neural network (skip network) is trained to predict end states from an arbitrary timepoint (time t-th in the depiction). Prediction errors are used to update weights of the skip network (yellow arrow).

*How are event schemas represented and learned?*

**Sparsely changing latent states network (GateL0RD):** This model uses a module called GateL0RD (Gumbsch et al. 2021) to represent all event schemas. GateL0RD is a modified RNN such that latent states change sparsely in time. GateL0RD imposes this inductive bias in the RNN by penalizing changes in the hidden layer vector in the loss function, in addition to prediction errors. Intuitively, the network learns to update its representations of latent factors only when necessary—when the cost of updating is outweighed by prediction accuracy. Importantly, two gates of GateL0RD are differentiable and thus learnable by gradient propagation: it learns to detect when there is a change in latent factors in the world and adjust its representation of these latent factors timely. These moments are signaled by patterns of activity in the hidden layer and the observed sensory input. The motivation for this modified RNN (GateL0RD) is the observation that many generative latent factors in the physical world are constant over extended periods of time, thus there's no need to update at every timestep. This hypothesis is similar to a hypothesis in causality research, arguing that latent factors interact sparsely in time and locally in space (Seitzer, Schölkopf, and Martius 2021). For example, consider a robotic arm in front of an object on a table. Clearly, the object can only be moved when contact between the robot and object is made (locally in space). However, there is no causal dependency between the arm and the object most of the time while the arm is moving (sparsely in time).

**Hierarchical structure and information flow (top-down, bottom-up, and lateral dynamics):** There are two levels of representations in the Gumbsch model. Sensory stimuli are represented at level 1, and latent states are represented at level 2 (fig. 5A). Level 2 latent states generate top-down sensory predictions at level 1 (fig. 5B, Predicting step). In this two-level model, level 1 to level 2 bottom-up propagations contains only representations. Level 1 representations are used to propose new level 2 representations of latent states and also to gate representation updating (fig. 5B, Updating step). Like other RNN variants, in this model level 1 representations in the previous timestep don't influence level 1 representations in the current timestep. Level 2 latent states in the previous timestep influence latent states in the current timestep via W_hidden. However, this W_hidden is set to an identity matrix by GateL0RD module if it decides that opening the gate is not necessary.

**Event schema learning:** in the Gumbsch model, all event schemas are represented by the three weight matrices connecting levels. What differentiates schemas are patterns of activity in the hidden layer vector. Different activity patterns in the hidden layer vector make different predictions about the upcoming location output given the same location input. The Gumbsch predicts current locations of object and robotic arm given current latent states (represented in the hidden layer vector) and the previous locations of the object and the arm. Observations are compared to predictions, and errors are propagated back to adjust weight matrices (W_out, W_in, W_hidden) (fig. 5C).

**End state learning**: In Kuperberg's proposal, the agent needs to learn a range of "end states" associated with each type so that it could proactively disengage the current event model and start searching the environment for cues about the next event model to construct**.** (Gumbsch et al. 2022) provides a computational solution to that proposal. This architecture adds a network that learns to predict the situations in which the latent states tend to change (skip network in fig. 5D), in addition to a network that learns to predict sensory input from hidden units (latent states). Event boundaries are defined as timepoints where hidden layer vector changes. Model-generated event boundaries (where latent states change) are used to generate end states, which are used to train the skip network (fig. 5D).

*When and how are working event models updated?*

In the Gumbsch model, working event models are represented by the hidden layer vector. Hidden layer vector is stable most of the time and updated when the gates are open. The gates are a function of prior activations of hidden units and current sensory information (fig. 5B, Updating step). Timepoints where GateL0RD changes its hidden layer vector were conceptualized as event boundaries.

*How selective attention controlled?*

Selective attention refers to the process of focusing on relevant sensory features. What is not represented in fig. 5B and 5D is that there are both uncertainty over prediction of next sensory information (blue square p in fig. 5B) and uncertainty over prediction of the end state (yellow square p in fig 5D) of the current event. Specifically, the skip network is trained to predict the distribution of final observation (instead of a point estimate) of an event from an arbitrary timepoint within event, and the forward network is trained to predict the distribution over next sensory observation. Intra-event uncertainty is modeled as predictive variance of the next timestep, in the forward network. Inter-event uncertainty is modeled as predictive variance of the end state of the current event, in the skip network. Assuming sensory input is always noisy, a gaussian noise is applied on all input vectors representing sensory input. The authors implemented selective attention as a noise mask on the observed input data. When the model directs its attention to a specific entity, such as the hand of an agent, all the related dimensions in the input data (e.g. the hand's position) remain free of sensory noise. In other words, the system can zero in on specific entities, similar to the way we would look at something, thereby receiving clear sensory data, while all other non-focused entities are subject to sensory noise. Critically, the entity that the system decides to attend to at a given timepoint depends on both intra-event uncertainty and inter-event uncertainty associated with that entity.

## A model of event dynamics and cooccurrences

The Elman and McRae model was introduced as a model of event knowledge—a model that learns co-occurrence patterns among components of events, and event temporal dynamics. The model views a sequence of event, each event comprises of a sequence of scenes (e.g. "going to a restaurant event" or "fixing a flat tire" event). Importantly, transitions between events are hand-coded by the experimenter and the model does not need to infer (whereas SEM, REPRISE, or Gumbsch models need to infer these transitions). Each scene is represented by a set of symbols (e.g. agent=John, action=go, location=restaurant, etc.). At a given timestep, the model is tasked with predicting the symbols in the next timestep, given previous symbols. To learns to predict the symbols in the next timestep, the model needs to learn the temporal dynamics of the event (trajectories of symbols). Moreover, in another task, the input symbols are not complete, and the model is tasked with inferring missing symbols—elaborative inference. To perform elaborative inference effectively, the models need to learn cooccurrence patterns among symbols that comprise an event.

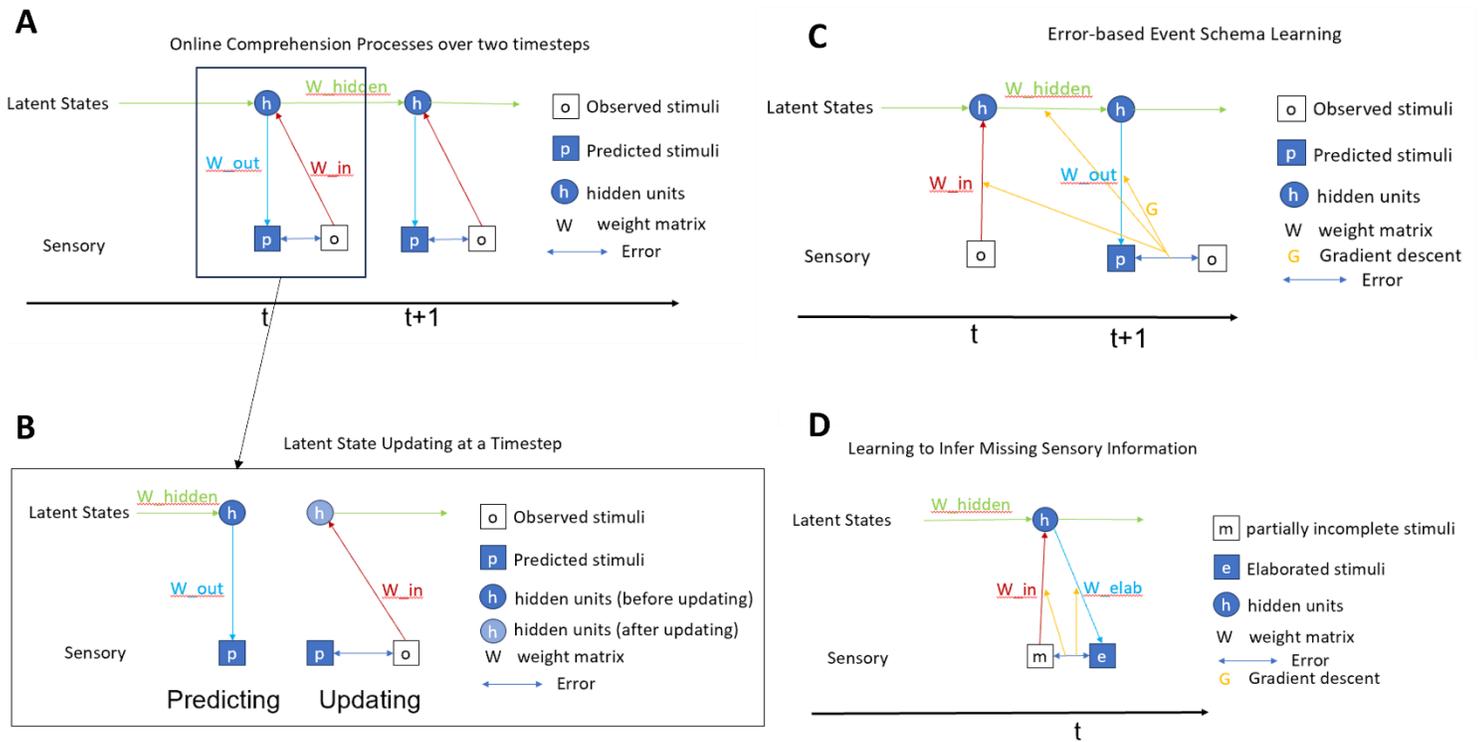

Figure 6: Mapping the two-level Elman and McRae model of event knowledge the the placeholder framework. A) Online comprehension processes over two timesteps. Feedback is represented by matrix W_out, lateral dynamics are represented by matrix W_hidden, feedforward are represented by matrix W_in. B) The two-step process: predicting sensory information and updating hidden layer vector. In Predicting step, activations of the hidden layer in the previous timestep affect hidden layer activations (level 2) of the current timestep (lateral dynamics in green arrow), which generates sensory prediction (level 1). In Updating step, only sensory observations (not errors) are propagated upward to influence hidden layer vector (level 2). C) Learning of event schemas: The learning involves making future predictions and adjusting the W_in, W_out, and W_hidden weight matrices according to prediction errors. D) Learning to infer missing information: the process of constraint satisfaction is implemented through the bidirectional connectivity between the sensory input and hidden layers. When the model encounters inputs that are only partially observable, these triggers specific activation patterns in the hidden layer. These patterns, which are shaped by previously learned co-occurrence statistics embedded within the bidirectional connections, then interact back with the input layer. This interaction serves to infer and activate components of the event that are missing or relevant in the input layer, effectively filling in gaps in the sensory information.

*How are event schemas represented and learned?*

**Hierarchical structure and information flow (top-down, bottom-up, and lateral dynamics):** Elman and McRae model is a modified RNN with an additional weight matrix for elaborative inference. There are two levels of representations: symbols are represented at level 1, latent states of the current event, or working memory, are represented at level 2 (fig. 6A). There are two types of level 2 to level 1 top-down projections. Level 2 representations in working memory project top-down predictions to level 1 (symbols in the next timestep) via W_out, this is typical in an RNN network (fig. 6B, Predicting step). In addition, Level 2 representations also project top-down elaborations to level 1 via W_elab to infer missing symbols in the current timestep (fig. 6D). Level 1 to level 2 propagations only contain representations (W_in) to update latent states in working memory (fig. 6B, Updating step). The W_hidden weight matrix defines how the content of working memory in the previous timestep influence the content of working memory in the current timestep—level 2 lateral dynamics. Like other RNN variants, in this model level 1 representations in the previous timestep don't influence level 1 representations in the current timestep.

**Event schema learning:** The model encodes all event schemas by a single RNN. RNNs are trained by sensory prediction errors through back propagation (fig. 6C).

*How components that comprise an activity are learned?*
One feature of human event (and discourse) comprehension is the ability to fill in missing information (elaborative inferences) from partially observable scenes. This can be achieved by learning the components that comprise an event, from co-occurrence statistics (e.g., instrument=cello is often associated with location=theatre-hall)**.** Elman and McRae model of event knowledge addresses this issue. Activities are comprised of various participants, actions, and contexts. The model is tasked to learn which components occur and co-occur across events. To learn co-occurrence statistics, there is a two-way connection between hidden units and input units (different weights connecting input units to hidden units, blue and red arrows in fig. 6D); this is a modification from a vanilla RNN. The two-way connection between input layer and hidden layer confers the model constraint satisfaction property, which aid elaborative inference. Constraint satisfaction refers to the process wherein the activation of a specific concept or idea is influenced by the degree of activation it receives from other related concepts and ideas. Essentially, when we think of one idea, it might trigger or reinforce the activation of another idea, especially if they are related or have been previously associated. This dynamic ensures that our mental representations are not random but are bounded by experiences. In other word, our mental representations depend on both by the immediate context (the activated concepts in the input) and by pre-existing knowledge of cooccurrences stored in our long-term memory. In the context of the Elman and McRae model, constraint satisfaction is exemplified by the two-way connection between the input layer and the hidden layer. When there's partially observable input, specific patterns are activated in the hidden layer. These patterns, based on learned co-occurrence statistics (embedded in the two-way connections), then reciprocally activate missing or related event components in the input layer.

*When and how are working event models updated?*
In this work, the authors didn't model event boundaries. Hidden layer vector, which represents the content of working event model, is updated continually as new sensory information comes in.

# 5. Comprehension processes give rise to hierarchical structured representations

## How event schemas are represented and learned
Schematic knowledge about events specifies how an event might unfold, and a range of possible end states. The human event comprehension system needs to learn various capacities to achieve that end. Schematic knowledge encodes how information in the space of modality-specific perceptual details used to construct higher-level multimodal representations where changes are smoother and slower. Schematic knowledge encodes how multimodal representations might evolve over time. Schematic knowledge also encodes how multimodal representations exert predictions of upcoming perceptual details. Event comprehension models propose mechanisms to achieve those computational goals. Thus, to compare how different models represent and learn event schemas, we need to compare their assumptions regarding: 1) hierarchical structure, 2) top-down influences, 3) bottom-up propagations, and 4) lateral dynamics within each level, 5) event schema learning, and 6) end states learning.

## Hierarchical Structure

In the SEM and REPRISE models, event representations are organized across three distinct levels. The first level, the scene level, is essentially the input/output of recurrent neural networks (RNNs). This level captures the immediate features of an event. The second level, referred to as the latent states, is represented by the hidden units within the RNNs. These hidden units capture underlying factors that are not be directly observable but play a critical role in influencing the scenes. Finally, the third level represents event types. In REPRISE (Butz et al., 2019), event type is defined by a vector while in SEM event type is defined by a categorical variable. This difference has implications in how top-down influence is assumed in the two models, described in the next section.

The Lu model (Lu et al. 2022), the Gumbsch model (Gumbsch et al. 2022) and the Elman and McRae (Elman and McRae 2019) model implemented two-level architectures: level 1 representing observable scenes, level 2 representing latent states of scenes using hidden units of an RNN. These models do not dedicate specific components to represent event types. The reason that the Lu model and the Elman and McRae model do not have representations of event types is that modelers are not interested in how or when humans detect changes in event types—event boundaries. In the Gumbsch model, event boundaries are conceptualized as changes in the hidden layer vector, whereas, in SEM and REPRISE, event boundaries are conceptualized as changes in the context vector.

In the five models, one general theme is that hierarchy is an inductive bias that modelers embed in their models. Moreover, the number of hierarchies is a hyper-parameter. However, one could assume that the shape of the hierarchy that a model comes to represent depends on the hierarchical structure of the environment. This raises two important questions for future modeling work: Is there a way to non-parametrically learn the number of hierarchical levels instead of treating it as a hyper-parameter? Relatedly, what are the mechanisms to spawn a new level? After a new level is spawn, one also need to specify top-down, bottom-up, and lateral dynamics. One approach is to first learn one level of representation and then gradually chunk these units into higher level representation, as demonstrated by the Hierarchical Chunking Model (Wu et al. 2021). Hierarchical Chunking Model works like hierarchical clustering algorithms, and thus it requires a hyper-parameter to decide when and how to group units into larger chunks.

## Mechanisms for top-down influence

SEM and REPRISE proposed three-level models while the Lu, the Gumbsch, and the Elman and McRae models proposed two-level models. In REPRISE, between context layer (level 3) and hidden layer (level 2), representations at the higher level (context layer) is propagated down, through feedback connections, to modulate activity at the lower level (hidden layer). In SEM, because each event schema is defined as a separate set of RNN weights (W_hidden, W_out, W_in), hidden layer vector is influenced by event type (via W_hidden). Moreover, top-down influence from hidden layer to sensory layer is also modulated by event type (via W_out) in SEM. In other four models, sensory predictions at level 1 are determined solely by the hidden layer vector at level 2. The common assumption among models is that sensory prediction of the next timestep is a function of latent states and event types, which is also the assumption in latent cause inference models (Gershman et al. 2014; Shin and DuBrow 2021).

## Mechanisms for bottom-up propagation

In a hierarchy, representations of a higher level can be constructed through two means. The first way is to construct increasingly abstract feature hierarchies, by using larger input context later in the hierarchy; this approach is typically implemented in convolutional networks, where units in a higher level have a larger receptive field and compute non-linear transformations of lower units, or recurrent neural networks, where units in a higher level serves as a low-dimensional information compression of the lower level. In this view, the feedforward information includes representations of the lower level. The second way is to infer representations of a higher level that best explain representations in a lower level, based on learned knowledge. The learned knowledge encodes a generative model to reconstruct representations of the lower level given activations of the higher level. Reconstructed representations are compared against observed representations at the lower level, and prediction errors are propagated upward (feedforward) to adjust higher level representations. In this view, feedback connections carry representations, but feedforward connections carry prediction errors.

In all five models, between sensory layer (level 1) and hidden layer (level 2), sensory observation is propagated upward to hidden layer through feedforward connections to construct representations at level 2. In REPRISE, sensory prediction errors are propagated upward to both hidden layer (level 2) and context layer (level 3), through retrospective neural inference—updating unit activations of hidden layer vector and context vector based on errors. In SEM, prediction errors at level 1 do not influence level 2 representations (hidden layer vector) but could directly influence representations at level 3: if sensory prediction errors at level 1 is high, the models will infer a different event type (level 3). This comparison highlights that SEM and REPRISE models implicitly adopt the view that both representations and errors are propagated upward, whereas the other three models implicitly assume that only representations are propagated upward.

Event segmentation theory proposed that perceptual information is gated and could only get through the gate after event boundaries. Only in the Gumbsch model, sensory observation is gated by a learnable module (GateL0RD). In addition, the Gumbsch model also adds a latent gate, assuming that the content of event models are constant most of the time. Future work might add perceptual or latent gates to other models to evaluate these hypotheses.

## Mechanisms for lateral dynamics at each hierarchical level

All five models all use variants of a recurrent neural network. In a typical recurrent neural network, previous scenes (level 1) do not directly influence the current scene. However, previous hidden unit activations (level 2) directly influence the current hidden unit activations (via W_hidden)—this defines how working event models evolve over time. These features of the typical RNNs are true for all models.

In REPRISE, previous event type representations (level 3) are copied directly to the current event type representations. In SEM, event type dynamics is governed by sticky-Chinese Restaurant Process, which assigns a high probability on the previous event type relative to the possibility of switching to a different event type (fig. 3A, Bayesian Inference module); these are hyper-parameters in the model. In short, level 3 lateral dynamics are not learned but are priors in the two models. One can imagine extend these models by specifying learning rules at this level. For REPRISE, since the representation at level 3 is a context vector, it is reasonable to impose recurrent connections among context vectors across time, parallel to the level 2 connections among hidden unit vectors. Inductive biases about timescales of updating at each level can be embedded by adding penalization terms into loss functions, an approach that was used in GateL0RD. For SEM, since representation at level 3 is a categorical variable and not a vector, learning lateral dynamics across event types can be implemented by a learnable Markov transition matrix. This approach might provide a better prior than the sticky Chinese Restaurant Process.

### How event schemas are learned

One common component among the five models is that they use variants of an RNN, and even though the hierarchical structures are different among models, as described above, event dynamics are encoded in weight matrices. All models learn these weight matrices by predicting the upcoming sensory stimuli, and back-propagate prediction errors to adjust weights. One notable difference is that for SEM, REPRISE, and GateL0RD, the models need to infer when there is a transition between events so that they can activate relevant schemas and/or update their internal representations. The Lu model and the Elman and McRae model do not need to solve this challenge, there is only one schema in the Lu model and the transitions are given in the Elman and McRae model.

For models that learn dynamics for more than one event types, they differ mainly on how shared and specific event dynamics are learned for all event types. REPRISE and GateL0RD have a single RNN to learn dynamics for all event types, whereas SEM encode dynamics of each event type in a single RNN. Sharing the same RNN means that event dynamics across event types are learned jointly in REPRISE and GateL0RD, thus the RNN can benefit from shared dynamics across event types. Having separate RNNs deprives SEM of the ability to learn shared dynamics across event types. In SEM-2.0 (Bezdek el al., 2023), the authors added a mechanism to initialize a new RNN based on a generic RNN trained on all stimuli observed thus far, effectively conferring the generic dynamics of the environment on the newly created schema. Having a shared RNN helps learn regularities across event types, but might also lead to interference (Tetko et al. 2019; Yu et al. 2020). This comparison between SEM and REPRISE and GateL0RD highlights different computational strategies to model the same cognitive mechanism, which is learning shared dynamics across event schemas. The models make different predictions: in SEM, if an event schema is already created, future learnings of other event schemas do not affect that event schema, whereas in REPRISE and GateL0RD, future learnings affect all event schemas.

### Mechanisms to learn end states at each hierarchical level

End states have a special status because they mark the completion of a goal-directed event, prompting observers to proactively disengage the current event model and prepare to construct the next event model (Kuperberg 2021). Moreover, end states can contain potential cues for the next event. For example, as we see a person reaching to a cup of coffee, depending on whether they grasp the cup handle or the spoon in the cup, the next event would be drinking coffee or stirring coffee. In the Gumbsch model, there are two levels of prediction: A skip network predicts the sensory state at the end of an event and a forward network predicts the next sensory state. Event boundaries, which are timepoints where hidden states change, are used to identify end states, which are used to train the skip network. The skip network is trained to predict the final observation of an event from an arbitrary timepoint within event.

The Gumbsch model defines end states as timepoints where the latent states (represented by the hidden layer vector) tend to change. Alternatively, instead of relying on patterns of latent states to identify end states, the model could also rely on prediction quality to identify end states. Two candidate measures of prediction quality are prediction error and prediction uncertainty. The intuition is that at the end of an event, prediction error and/or prediction uncertainty tend to rise.

In short, learning end states is an important computational challenge in modeling event comprehension. To learn end states in an unsupervised manner (no external signals of when an event ends and another begins), there are at least two mechanisms to self-generate end states: 1) to rely on moments when the content of working event model shift, which was proposed by the Gumbsch model and 2) to rely on moments when prediction quality degrades, which has not been tested in any model.

### When and how event models are updated

In the REPRISE model, working event models are represented by two primary components: the context vector and the hidden layer vector. Key to this model is the continuous, un-gated adjustment of both the hidden layer vector and the context vector through the feedback mechanism of prediction errors.

The SEM model uses the hidden layer vector and the event type categorical variable to represent working event models. Different from REPRISE, in SEM updates to the event type is punctuated: event types remain the same most of the times and are likely to change when prediction errors are high. Like REPRISE, updates to the event type are influenced by prediction errors. Updates to the hidden layer vector are continual: small drifts to incorporate new sensory information when consecutive timesteps share the same event type, and dramatic shifts to establish a new working event model when there is a change in event types.

The Gumbsch model introduces a unique mechanism: a gate that controls the stability and updating of the hidden layer vector, which also represent the working event model. This gate is influenced by prior activations of the hidden layer and current sensory information.

The Lu Model is primarily interested in the role of episodic memory in prediction and only learn one event schema. The hidden layer vector, representing the working event model, is continually updated with sensory information. A distinctive feature of this model is the influence of stored episodic memories on the activations of the hidden layer vector, enabling a form of memory retrieval in the process.

In the Elman and McRae model, event boundaries are not explicitly modeled. The working event model, represented by the hidden layer vector, is continuously updated to incorporate new sensory information.

Each of these models offers a unique perspective on how working event models are updated, reflecting various theoretical underpinnings and computational strategies. REPRISE and SEM emphasize the role of prediction errors on event model updating, whereas the other three models do not rely on prediction errors to update the content of working event models. In other words, REPRISE and SEM implicitly assume that comprehension process works like generative models: hidden latent states generate observations, and errors are propagated upward to revise latent states. The Lu model integrates the influence of memory retrieval, assuming humans rely on episodic memories as schema-based prediction becomes difficult. The Gumbsch model emphasizes a learnable gating mechanism, which is complementary to error-based gating proposed by event segmentation theory (Zacks et al. 2007).

## Learning components comprising an event

Elaborative inference—the ability to fill in missing information in partially observable scenes—is crucial for understanding events or discourses. The Elman and McRae connectionist model of event knowledge (2019) learns co-occurrence patterns among components of activities, such as entities, actions, and contexts, by a feature of their network: constraint satisfaction. Specifically, the input units representing components of an event are fully connected to hidden units, and hidden units project recurrent connections to input units. The two-way connection between input layer and hidden layer confers the model constraint satisfaction property, which aids elaborative inference**:** "partially" complete input symbols activate a certain pattern in hidden units, which in turn refine activations of input symbols (inferring missing components), which in turn activate another pattern in the hidden units. These updating cycles continue until the activations in both input layer and hidden layer settle into stable patterns (hence constraint satisfaction).

The Elman and McRae models is the only event comprehension model trying to capture how a system learns to fill missing information. However, it's unclear how elaborative inference and prediction interact. For example, inferring missing information before making predictions might help increase prediction accuracy. Or, suppose that the system only make elaborative inference when necessary, timepoints where the system make elaborative inference could be when prediction quality degrades, or when predictions are abnormal under the distribution of observed sensory information. Future modeling efforts are needed to characterize prediction-elaboration interaction.

## Learning when to retrieve from memory

As discussed earlier, the Lu model proposes a computational model of event comprehension, suggesting that people selectively retrieve episodic memories (Chen et al. 2016). The researchers trained a memory-augmented neural network using reinforcement learning to predict a feature (e.g. mood) in an environment where features that inform the prediction might have been observed (e.g. weather). The network learned to decide whether it has observed the feature and answer or say "don't know" (uncertain about the feature). If it decides that it doesn't know the feature, it will retrieve memories based on their similarities with the current LSTM hidden unit representations and integrate them into LSTM hidden unit representations. In addition, the author invites us to think about other factors affecting whether we retrieve memory, such as familiarity signal pushed the model into "retrieval mode" while high between-event similarity environments made the model less likely to retrieve. The model predicts that people selectively encode at the end of the event instead of the middle of the event—the model's prediction accuracy is higher if it encodes end-event memories.

In this model, episodic memory encoding and retrieval are assumed to be lossless and static. These assumptions are reasonable given that the authors are interested in when to encode and retrieve memories. For researchers who are interested in lossy (presumably more realistic) encoding and retrieval processes, the Structured Event Memory (SEM) model offers an approach. Unlike traditional stationary models like the Hidden Markov Machine (HMM), SEM emphasizes the dynamic nature of events. The goal of both HMM and SEM is to reconstruct the original scene as close as possible, given a noisy memory trace. The two models compensate noisiness combining learned semantic knowledge with the noisy memory trace to reconstruct the original scene. This can be thought of as using semantic knowledge to regularize memory traces. What sets SEM apart from an HMM is the way it regularizes recalled memories. Whereas HMMs rely on central tendencies or "averages" of scenes within an event to regularize a recalled scene, SEM relies on the learned dynamics of the event to regularize recalled scenes. Stationary models like HMMs are suitable to model static memories such as memories for fruits and animals, whereas SEM is more suitable to model dynamic memories such as event memories.

Episodic memories are important to comprehend events or detect event changes (Stawarczyk et al. 2020; Wahlheim and Zacks 2019). Although existing models like the Lu model offer valuable insights into when and how memories are retrieved, there is still much to explore. For one, when memories should be encoded could be framed as a mechanistic solution to a computational challenge, such as reconstructing the events with few episodic memories. For another, memory about association between observations and latent causes can be used to efficiently infer latent causes given the current observation, as studied by (Lu et al. 2023).

## Mechanisms for attentional control

The Gumbsch model architecture allows for comparisons between three different versions of attention selection: Minimizing uncertainty within the current event (intra-event uncertainty), minimizing uncertainty about the next event boundary (inter-event uncertainty), or minimizing both uncertainties. Recall that the model has two networks: the forward network and the skip network. Intra-event uncertainty is modeled as prediction uncertainty over the next timestep, in the forward network. Inter-event uncertainty is modeled as prediction uncertainty over the end state of the current event, in the skip network. Importantly, the entity that the system decides to attend to at a given timepoint depends on the degree of both intra-event uncertainty and inter-event uncertainty associated with that entity.

When only reducing intra-event uncertainty, the model primarily focused on the hand well before the hand made contact with the object. Focusing on the hand in this way aids in forecasting immediate hand movements as it reaches for the object. In the case of minimizing only inter-event uncertainty, the system generally directed its attention first to the object before the hand reached it. It appears that the system has learned that concentrating on the object helps to reduce uncertainty regarding when and where the reaching event will conclude, and which future events will follow (e.g. lift the object). Crucially, when combining both types of uncertainties, the system displays attention shifts that are similar to 7- to 11-month-old infants. In the event sequence of reaching, grasping, and transporting, the system generally first focused on the hand, and then shifted its attention to the object before the hand made contact with the object. These patterns are similar to the gaze behaviors observed in infants while they watch videos of a hand grabbing and lifting a toy, as studied by (Adam and Elsner 2020). At 6 months old, infants mainly track the hand with their gaze. However, older infants, at 7 and 11 months, transition their gaze from the hand to the object before the hand reaches it. These results make an interesting prediction: During online comprehension, the level of representation that is most relevant to the observer will control where the observer looks to minimize uncertainty over representations at that level.

In the realm of episodic memory control, uncertainty plays a pivotal role in shaping how the network utilizes its stored memories for future predictions (Lu et al. 2022). As discussed, the network employs a gating mechanism that adjusts the influence of episodic memories on the hidden units. This adjustment is guided by the current state of the network's working memory, also represented by hidden units. For instance, if the network is tasked with predicting emotional states like happiness or anger but hasn't yet received data on the weather, a known influencing factor, it adjusts its episodic memory gate to be more open. The level of this gate's openness is determined by the current state of the network's working memory, which in this case, lacks information about the weather. The lack of information about the weather has a signature: When faced with insufficient data to make a specific prediction, the ac'Ivity of hidden units is reduced. This reduced activity is interpreted as the network being"""uncertain""" leading it to produce an""""unknow""" choice as its output.

This mechanism of episodic memory control based on uncertainty complements the attentional control mechanism described by Gumbsch et al., 2022. Both mechanisms aim to minimize uncertainty, but they do so in different domains: attentional control focuses on sensory input, while episodic memory control focuses on leveraging past experiences stored in memory.

## Constructing structured representations

Thus far, we have been concerned with processes aiding construction and updating of event models, but we have not talked about the content of event models. Here, because we are focusing on events involving human activities, it is important to consider which features could be useful to comprehend this type of events. The Event Indexing Model proposes that humans track five dimensions: time, space, causality, intentionality, and protagonist. The Event Horizon Model proposes the content of event models includes spatiotemporal frameworks, entities and their properties, structural relations and linking relations (Radvansky and Zacks 2014). Some examples of structural relations are spatial relations, ownership relations, kinship relations, social relations, and so forth. Linking relations serve to link different events into a sequence. The most common types of linking relations are temporal and causal relations. Temporal relations often covary with causal relations because causes always precede their effects. However, causal relations are usually much more important because they license predictions about the future. Both theories emphasize representations of not just entities involved in activities but their structural and linking relations, especially causal relations. Propositional representation is a good medium to capture entities, their properties, and relations among entities (Kintsch 1998, 2001). The computational models reviewed so far either give relational information directly to the model (e.g., Elman and McRae model, SEM model with Holographic Reduced Representation) or operate in a constructed world where relational information is not critical to comprehend events (The Gumbsch model, the Lu model, and REPRISE model). Moreover, there is an implicit assumption that the RNN (or its variants such as long short-term memory or gated recurrent unit) might be able to extract underlying structural relations and temporal relations among entities to be able to predict how the event will unfold. This assumption needs to be tested if modelers of event comprehension want to use these networks to model dynamics governing naturalistic events. Alternatively, we can rely on advances in computer vision methods to automatically extract entities and their structural relations on the scene (Ji et al. 2020) and use symbolic techniques such as Holographic Reduced Representations (Plate 1995, 1997) or Tensor Products (Smolensky 1994; Smolensky et al. 2016) to translate this information into a continuous vector space as input to connectionist architectures such as RNNs. Another approach is to augment an RNN with modules that are designed to progressively extract abstract representations such as in convolutional neural networks. (Lotter, Kreiman, and Cox 2017, 2018) used convolutional LSTMs (cLSTMs) as representation modules for the task of next frame video prediction. A convolutional LSTM (cLSTM) is an enhanced version of the standard LSTM neural network, designed specifically for image sequence processing like in video analysis. While traditional LSTMs are effective for tasks that involve sequences, such as natural language processing, they use feature vectors as their internal data structure. In contrast, cLSTMs use multi-channel images, making them better suited for capturing spatial patterns in image data. This change enables cLSTMs to capture both spatial and temporal correlations more effectively. Experiments have shown that cLSTMs consistently outperform traditional LSTMs in tasks that involve spatiotemporal data (SHI et al. 2015). However, it's unclear whether cLSTMs could capture the full range of structural and temporal relations at the scale of human naturalistic activities.

Relatedly, in exploring how one can generalize learned structural and temporal relations to new situations, it is crucial to understand the concept of role-filler bindings. Consider, for instance, frequenting a coffee shop. Each visit is unique in its details, yet they share a schematic structure that we can rely on to comprehend new experiences. In the context of a coffee shop, the schema consists of roles including the barista, the drink, and the customer. These roles act as placeholders that can be filled by specific entities—known as fillers—such as Alice, tea, or Bob. Therefore, when we hear a statement like "Alice ordered a tea from Bob," we automatically link Alice to the role of customer, tea to the role of drink, and Bob to the role of barista, even if we're unfamiliar with the specific terms Alice, tea, or Bob (Chen et al. 2021). The mechanism that allows us to make these connections is known as role-filler binding. It is the cognitive operation that links specific fillers to general roles, forming the basis for constructing propositional representations. Chen et al. examined whether RNNs or their variants can learn role-filler bindings. Among the five reviewed models, role-filler binding is either the input to the models (localist symbolic input in Elman and McRae model, Holographic Reduced Representation in SEM model) or assumed to be learned by the RNN. (Chen et al. 2021) tested four model architectures, focusing on how different networks learn to solve the task of binding novel fillers to learned roles. This study demonstrated that RNN or LSTM were inadequate at the task, while networks with external memory (fast weights RNN (Ba et al. 2016) and Differentiable Neural Computer (Graves et al. 2016)) can learn to bind roles to arbitrary fillers without explicitly labeled role-filler pairs much better.

In summary, current methods in modeling human event comprehension are limited in their ability to adequately represent structured information. Binding roles to fillers is a crucial step in constructing structured representations. Traditional Recurrent Neural Networks (RNNs) and their variants like Long Short-Term Memory networks (LSTMs) fall short in automatically learning role-filler binding operations in the text domain. Given that LSTMs are designed to process texts and that they fall short on processing image sequences, where role filler binding in visual domain requires extracting more abstract concepts (e.g. objects, actors, etc.), LSTMs should not be expected to perform role-filler binding well for image sequences. To address these limitations, a more sophisticated approach is necessary, especially in modeling projects where learning propositional representation is critical. One potential way forward is to integrate advanced techniques that can enhance the existing architectures. For instance, incorporating models with external memory capabilities, such as Fast Weights RNNs or Differentiable Neural Computers, has shown promise in learning role-filler bindings more effectively in the text domain. Furthermore, augmenting RNNs with specialized modules, such as convolutional LSTMs, can significantly improve the handling of spatiotemporal data, thereby enhancing the model's capacity to adequately represent and process the complex structural and temporal dependencies. This multi-faceted approach, integrating various advanced techniques, presents a promising direction for overcoming the current limitations and advancing the field of event comprehension modeling.

## 6. Conclusions and Future Directions

The comparison of computational models (Butz et al. 2019; Elman and McRae 2019; Franklin et al. 2020; Gumbsch et al. 2022; Lu et al. 2022) of event comprehension has revealed several key themes and areas for further research:

**Hierarchical processing as inductive bias**: A common thread across the models is the use of hierarchical structures as an inductive bias. This reflects a fundamental aspect of human cognition, where understanding events involves processing information at multiple levels, from immediate sensory perception to more abstract conceptualization. However, a significant area for future research lies in the potential for these models to not only utilize predefined hierarchical levels but also to learn and construct new levels within the hierarchy autonomously. Exploring mechanisms that enable a model to dynamically generate and incorporate additional hierarchical levels could lead to adaptive systems suitable for different types of discourses or events. Such development could involve learning algorithms that identify when existing hierarchical structures are insufficient and initiate the creation of new levels or the reorganization of existing ones.

**Sensory prediction**: The five models generally agree that sensory prediction of the next timestep is a function of latent states and event types. This is aligned with assumptions of latent cause models, suggesting that understanding events involves inferring underlying causes from sensory inputs.

**Propagation of representations and errors**: A key distinction emerges in how models handle bottom-up propagation. SEM and REPRISE suggest that both representations and prediction errors propagate upward through the hierarchy. In contrast, the other models assume upward propagation of representations alone. Teasing apart the contributions of each flow of information in the computational models is an important step toward characterizing the flow of information in human event comprehension since it not only informs potential mechanisms but also provides heuristics for experimental research.

**Event segmentation and gating mechanisms**: The Gumbsch model gates perceptual information. Gating perceptual input was also introduced in a computational model of event segmentation not reviewed here (Reynolds, Zacks, and Braver 2007) (I should probably add Reynolds's model when this is turned to a manuscript). This approach is consistent with event segmentation theory and suggest the feasibility for incorporating gating mechanisms in other models to assess their impact on model performance. The gates introduced in the Gumbsch and Reynolds models are binary gates--either open or closed. Another potential gating mechanism is continuous gates. One implementation of continuous gates is through Bayesian updating: the degree of gate opening depends on the degree of uncertainty over latent states (Kuperberg 2021). This involves modeling uncertainties over hidden states. Modeling uncertainties in deep neural networks has long been a challenge due to its computational cost. Theoretically, one would need to train multiple deep neural networks to estimate variance associated with each weight, and this training requirement is a bottleneck in the field of Bayesian deep learning. However, recent advances have been made so that variances can be approximated from a single deep neural network (Gal and Ghahramani 2016; Kendall and Gal 2017). This opens a venue for future research interested uncertainty.

**Diverse strategies to learn shared dynamics**: A comparison between models like SEM, REPRISE, and GateL0RD reveals diverse strategies in modeling shared dynamics across event schemas. This range of strategies suggest potential cognitive mechanisms that could solve the same computational challenge.

**Learning lateral dynamics of event types**: The lateral dynamics at higher hierarchical levels, particularly in SEM and REPRISE, are currently treated as priors. However, there's a potential to enhance these models by integrating learning mechanisms at this level. For the REPRISE model, where the third level is represented by a context vector, it's feasible to establish recurrent connections among these context vectors over time. Incorporating inductive biases related to the timescales of updates at each level could further refine this process, perhaps through the inclusion of penalization in the loss functions, a method utilized in the GateL0RD model. In contrast, the SEM model uses categorical variables for its third-level representation. Here, learning the lateral dynamics across event types could be effectively achieved by developing a Markov transition matrix. This approach helps the model to extract statistical regularities in the data and potentially provides a more accurate prior than the currently used sticky Chinese Restaurant Process.

**Learning end states**: Learning end states remains a significant computational challenge. The Gumbsch model proposes relying on shifts in working event models, while the potential of using moments of degraded prediction quality as a cue for end states remains unexplored.

**Updating the content of working event models**: REPRISE and SEM emphasize the influence of prediction errors in updating event models, whereas the Lu, Gumbsch, and Elman and McRae models do not rely on these errors for updates. In addition, the incorporation of memory retrieval mechanisms, as seen in the Lu model, and gating mechanisms, as in the Gumbsch model, demonstrate the varied approaches to controlling the content of working event models. A promising direction is the integration of multiple mechanisms, like prediction errors, memory retrieval, and gating, within a single model. Future models integrating multiple mechanisms could benefit from more adaptive updating algorithms that can adjust their approach. For instance, a model might rely more on sensory information in unfamiliar situations but switch to memory-driven updates in familiar contexts.

**Interaction between elaborative inference and prediction**: The relationship between elaborative inference (filling in missing information) and prediction is yet to be clearly understood. One potential direction is exploring the neural underpinnings of elaborative inference and prediction to gain algorithmic insights that could be tested and studied in computational models.

**Challenges in learning structured information**: Current models face limitations in adequately representing structured information, which is crucial for understanding complex human activities and interactions within events. These limitations are primarily due to inherent limitations of recurrent neural networks (RNNs). Although convolutional LSTMs (cLSTMs) are a step forward in capturing temporal and spatial relations, more complex architectures may be required for capturing other types of relations such as causal relations, which has been emphasized by both discourse and event comprehension theories.

**Minimizing uncertainty with different means**: Theories of event comprehension suggest that humans are perpetually engaged in forecasting future events. These predictions, inherently imperfect, are always accompanied by a certain degree of uncertainty. In addressing this uncertainty, the Gumbsch and Lu models offer distinct but complementary strategies. The Gumbsch model focuses on reducing uncertainty through selective attention to sensory input, effectively prioritizing information that is most relevant for immediate comprehension and prediction. This model aligns with the idea that our sensory focus can be dynamically adjusted to enhance the clarity of task-relevant information. On the other hand, the Lu model emphasizes the role of episodic memory in managing uncertainty. By leveraging past experiences, this model allows for a more context-rich interpretation of events, where previous encounters inform current understanding and future predictions. Future developments may further explore how these mechanisms interact and can be optimally balanced or integrated in computational models to mimic human cognitive processes more accurately.

In the realm of computational modeling of event comprehension, a notable observation is the absence of a unified mechanistic model that addresses all the computational challenges. This scenario underscores the complexity and scale inherent in modeling human event comprehension. Each model, as reviewed, tackles specific aspects or challenges, indicating that the task at hand is multifaceted. Moving forward, there's a significant opportunity for advancement in this field by adopting a reductionist approach, akin to strategies used in the experimental domain. By focusing on enhancing our understanding of each individual component pertaining to event comprehension, we can lay a more solid foundation for future integration. This approach has demonstrated its efficacy in experimental settings, where breaking down complex phenomena into smaller, more manageable parts has led to deeper insights and more robust theoretical frameworks. In the context of computational modeling, this could mean refining aspects such as hierarchical processing, sensory prediction, and the learning of lateral dynamics separately before attempting to integrate them into a unified framework. Such a methodical and incremental approach is not only practical but may also prove to be instrumental in unraveling the complexities of event comprehension and ultimately leading to the development of a comprehensive, unified computational model.